\documentclass[runningheads]{llncs}
\usepackage{balance}
\usepackage{amsmath}
\usepackage[tight,footnotesize]{subfigure}
\usepackage{bm}
\usepackage{algorithm}
\usepackage[noend]{algorithmic}
\usepackage{multirow}
\usepackage{amsfonts}

\begin{document}




\title{Group Fairness in Non-monotone Submodular Maximization}
%
%
\author{Jing Yuan\inst{1} \orcidID{0000-0001-6407-834X} \and Shaojie Tang\inst{2}  \orcidID{0000-0001-9261-5210}}
\authorrunning{J. Yuan and S. Tang}
%
\institute{Department of Computer Science and Engineering, University of North Texas \\\email{jing.yuan@unt.edu} \and Naveen Jindal School of Management, University of Texas at Dallas\\\email{shaojie.tang@utdallas.edu} }

\maketitle              
\begin{abstract}Maximizing a submodular function has a wide range of applications in  machine learning and data mining. One such application is data summarization whose goal is to select a small set of representative and diverse data items from a large dataset. However, data items might have sensitive attributes such as race or gender, in this setting, it is important to design \emph{fairness-aware} algorithms to mitigate potential algorithmic bias that may cause over- or under- representation of particular groups. Motivated by that, we propose and study the classic non-monotone submodular maximization problem subject to novel group fairness constraints. Our goal is to select a set of items that maximizes a non-monotone submodular function, while ensuring that the number of selected items from each group is proportionate to its size, to the extent specified by the decision maker. We develop the first constant-factor approximation algorithms for this problem. We also extend the basic model to incorporate an additional global size constraint on the total number of selected items.
\end{abstract}


\section{Introduction}
Submodular function refers to a broad class of functions which satisfy the natural diminishing returns property: adding an additional item to a larger existing subset is less beneficial. A wide range of machine learning and AI problems, including exemplar-based clustering \cite{dueck2007non}, feature selection \cite{das2008algorithms},  active learning \cite{golovin2011adaptive}, influence maximization in social networks \cite{tang2020influence}, recommender system \cite{el2011beyond}, and diverse data summarization \cite{sipos2012temporal}, can be formulated as a submodular maximization problem. This problem, whose goal is to select a set of items to maximize a submodular function, and its variants \cite{gu2022binary,shi2021k} have been extensively studied in the literature subject to various constraints, including cardinality, matroid, or knapsack-type restrictions.

 We notice that in practise, items or individuals are often associated with different groups based on various attributes, such as gender, race, age, religion, or other factors. Existing algorithms might exhibit bias if left unchecked, for example, some of the groups might be over- or under-represented in the final selected subset. Therefore, it becomes increasingly important to design \emph{fairness-aware} algorithms to mitigate such issues. Towards this end, we propose and study the classic non-monotone submodular maximization problem subject to novel group fairness constraints.
 Our goal is to select a \emph{balanced} set of items that maximizes a non-monotone submodular function, such that the ratio of selected items from each group to its size is within a desired range, as determined by the decision maker. Non-monotone submodular maximization has multiple compelling applications, such as feature selection \cite{das2008algorithms}, profit maximization \cite{tang2021adaptive}, maximum cut \cite{gotovos2015non} and data summarization \cite{mirzasoleiman2016fast}. Formally, we consider a set $V$ of items (e.g., datapoints) which are partitioned into $m$ groups: $V_1, V_2, \cdots, V_m$ such that items from the same group share same attributes (e.g., gender). We say that a set $S\subseteq V$ of items is $(\alpha,\beta)$-fair if for all groups $i \in[m]$, it holds that $\lfloor\alpha |V_i|\rfloor \leq |S \cap V_i| \leq \lfloor\beta |V_i|\rfloor$.  Using our model, it allows for the decision maker to specify the desired level of fairness by setting appropriate values of $\alpha$ and $\beta$. Specifically, setting $\alpha=\beta$ leads to the highest level of fairness in that the number of selected items is \emph{strictly} proportional to its group size; if we set $\alpha=0$ and $\beta=1$, there are no fairness constraints. Our goal is to find such a $(\alpha,\beta)$-fair subset of items that maximizes a submodular objective function. Our definition of fairness, which balances solutions with respect to sensitive attributes, has gained widespread acceptance in the academic community, as demonstrated by its frequent use in previous studies \cite{celis2018multiwinner,el2020fairness,chierichetti2019matroids}. There are several other notations of fairness that can be captured by our formulation such as the $80\%$-rule \cite{biddle2017adverse}, statistical parity \cite{10.1145/2090236.2090255} and proportional representation \cite{monroe1995fully}.

\subsection{Our Contributions}

\begin{itemize}
\item Our study breaks new ground by examining the classic (non-monotone) submodular maximization problem under $(\alpha,\beta)$-fairness constraints. Our model offers flexibility in capturing varying degrees of fairness as desired by the decision maker, by adjusting the values of $\alpha$ and $\beta$.
\item We develop the first constant-factor approximation algorithm for this problem. We observe that the parameter $\alpha$ is closely linked to the complexity of solving the $(\alpha,\beta)$-fair non-monotone submodular maximization problem. In particular, when $\alpha\leq 1/2$, we design a $\frac{\gamma}{2}$-approximation algorithm and when $\alpha> 1/2$, we develop a $\frac{\gamma}{3}$-approximation algorithm, where $\gamma$ is the approximation ratio of the current best algorithm for matroid-constrained submodular maximization. We also extend the basic model to incorporate an additional global size constraint on the total number of selected items. We provide approximation algorithms that have a constant-factor approximation ratio for this extended model.
\end{itemize}

\subsection{Additional Related Works}
In recent years, there has been a growing awareness of the importance of  fair and unbiased decision-making systems. This has led to an increased interest in the development of fair algorithms in a wide range of applications, including  influence maximization \cite{tsang2019group}, classification \cite{zafar2017fairness}, voting \cite{celis2018multiwinner}, bandit learning \cite{joseph2016fairness},  and data summarization \cite{celis2018fair}. Depending on the specific context and the type of bias that one is trying to mitigate, existing studies  adopt different metrics of fairness. This can lead to different optimization problems and different fair algorithms that are tailored to the specific requirements of the application. Our notation of fairness is general enough to capture many existing notations such as the $80\%$-rule \cite{biddle2017adverse}, statistical parity \cite{10.1145/2090236.2090255} and proportional representation \cite{monroe1995fully}.  Unlike most of existing studies on fair submodular maximization \cite{celis2018multiwinner} whose objective is to maximize a monotone submodular function, \cite{el2020fairness} develop fair algorithms in the context of streaming non-monotone submodular maximization. Their proposed notation of fairness is more general than ours, leading to a more challenging optimization problem which does not admit any constant-factor approximation algorithms. \cite{tang2023randomization,tang2023unified} aim to develop randomized algorithms that satisfy average fairness constraints. Very recently, \cite{tang2022group} extend the studies of fair algorithms to a more complicated adaptive setting and they propose a new metric of fairness called group equality.

\section{Preliminaries and Problem Statement}

We consider a set $V$ of $n$ items. There is
a non-negative submodular utility function $f: 2^V\rightarrow \mathbb{R}_+$. Denote by $f(e\mid S)$ the marginal utility of $e\in V$ on top of $S\subseteq V$, i.e., $f(e\mid S)=f(\{e\}\cup S)-f(S)$. We say a function  $f: 2^V\rightarrow \mathbb{R}_+$ is submodular if for any two sets $X, Y\subseteq V$ such that $X\subseteq Y$ and any item $e \in V\setminus Y$,
\[f(e\mid Y) \leq f(e\mid X).\]

Assume $V$ is partitioned into $m$ disjoint groups: $V_1, V_2,\cdots, V_m$. We assume that there is a given
lower and upper bound on the fraction of items of each group that must be contained in a feasible solution. These two bounds, namely $\alpha$ and $\beta$, represent group fairness constraints. The problem of $(\alpha, \beta)$-fair submodular maximization problem  (labelled as $\textbf{P.0}$) can be written as follows.

 \begin{center}
\framebox[0.5\textwidth][c]{
\enspace
\begin{minipage}[t]{0.5\textwidth}
\small
$\textbf{P.0}$
$\max f(S)$ \\
\textbf{subject to:}  \\
$\lfloor\alpha |V_i|\rfloor \leq |S \cap V_i| \leq \lfloor\beta |V_i|\rfloor, \forall i \in[m]$.
\end{minipage}
}
\end{center}
\vspace{0.1in}

One can adjust the degree of group fairness in a feasible solution through choosing appropriate values of $\alpha$ and $\beta$. I.e., strict group fairness is achieved at $\alpha=\beta$ in which case every feasible solution must contain the same $\alpha$ fraction of items from each group; if we set $\alpha=0$ and $\beta=1$, then there is no group fairness constraints. We next present the hardness result of this problem.

\begin{lemma}
Problem $\textbf{P.0}$ is NP-hard.
\end{lemma}
\emph{Proof:} We prove this through reduction to the classic \emph{cardinality constrained submodular maximization problem} which we define below.
\begin{definition} The input of  cardinality constrained submodular maximization problem is a group of items $U$, a submodular function $h: 2^U\rightarrow \mathbb{R}_+$, and a cardinality constraint $b$; we aim to select a group of items $S\subseteq U$ such that $h(S)$ is maximized and $|S|\leq b$.
\end{definition}

We next show a reduction from cardinality constrained submodular maximization problem to $\textbf{P.0}$.
Consider any given instance of cardinality constrained submodular maximization problem, we construct a corresponding instance of $\textbf{P.0}$ as follows: Let $V=U$, $f=h$, assume there is only one group, i.e., $V=V_1$, and let $\alpha=0$, $\beta=b/|U|$. It is easy to verify that these two instances are equivalent. This finishes the proof of the reduction. $\Box$

\section{Non-monotone Submodular Maximization with Group Fairness}
\label{sec:adaptive-general}
\paragraph{Warm-up: Monotone Utility Function} If $f$ is monotone and submodular, we can easily confirm that $\textbf{P.0}$ can be simplified to $\textbf{P.1}$ by removing the lower bound constraints. This is because in this case, increasing the size of a solution by adding more items will not decrease its utility. As a result, the lower bound constraints in $\textbf{P.0}$, which state that $\lfloor\alpha |V_i|\rfloor \leq |S \cap V_i|$ for all $i \in[m]$, can always be met by adding sufficient items to the solution.

 \begin{center}
\framebox[0.4\textwidth][c]{
\enspace
\begin{minipage}[t]{0.4\textwidth}
\small
$\textbf{P.1}$
$\max f(S)$ \\
\textbf{subject to:}  \\
$|S \cap V_i| \leq \lfloor\beta |V_i|\rfloor, \forall i \in[m]$.
\end{minipage}
}
\end{center}
\vspace{0.1in}

Since $f$ is a monotone submodular function, $\textbf{P.1}$ is a well-known problem of maximizing a monotone submodular function subject to matroid constraints\footnote{A matroid is a pair $\mathcal{M} = (V, \mathcal{I})$ where $\mathcal{I} \subseteq 2^V$ and
1. $\forall Y \in \mathcal{I}, X \subseteq Y \rightarrow X \in \mathcal{I}$.
2. $\forall X, Y \in \mathcal{I}; |X| < |Y| \rightarrow \exists e\in Y \setminus X; X \cup\{ e\} \in \mathcal{I}$.}. This problem has a $(1-1/e)$-approximation algorithm.

We then proceed to develop approximation algorithms for non-monotone functions. We will examine two scenarios, specifically when $\alpha\leq 1/2$ and when $\alpha> 1/2$.
\subsection{The case when $\alpha\leq 1/2$}
In the scenario where $\alpha\leq 1/2$, we use the solution of $\textbf{P.1}$ as a building block to construct our algorithm. First, it is easy to verify that $\textbf{P.1}$ is a relaxed version of  $\textbf{P.0}$ with lower bound constraints $\lfloor\alpha |V_i|\rfloor \leq |S \cap V_i|$ in $\textbf{P.0}$ being removed. Because $f$ is a submodular function, $\textbf{P.1}$ is a classic problem of maximizing a (non-monotone) submodular function subject to matroid constraints. There exist effective solutions for $\textbf{P.1}$. Now we are ready to present the design of our algorithm as below.
\begin{enumerate}
\item Apply the state-of-the-art algorithm $\mathcal{A}$ for matroid constrained submodular maximization to solve $\textbf{P.1}$ and obtain a solution $A^{P.1}$.

\item Note that $A^{P.1}$ is not necessarily a feasible solution to $\textbf{P.0}$ because it might violate the lower bound constraints $\lfloor\alpha |V_i|\rfloor \leq |S \cap V_i|$ for some groups. To make it feasible, we add additional items to $A^{P.1}$. Specifically, for each group $i\in[m]$ such that $|A^{P.1} \cap V_i|< \lfloor\alpha |V_i|\rfloor$, our algorithm selects a backup set $B_i$ of size $\lfloor\alpha |V_i|\rfloor-|A^{P.1} \cap V_i|$, by randomly sampling  $\lfloor\alpha |V_i|\rfloor-|A^{P.1} \cap V_i|$ items from $V_i\setminus A^{P.1}$. Define $B_i=\emptyset$ if $|A^{P.1} \cap V_i|\geq \lfloor\alpha |V_i|\rfloor$.
\item At the end, add $\cup_{i\in [m]}B_i$ to $A^{P.1}$ to build the final solution $A^{approx}$, i.e., $A^{approx}= A^{P.1}\cup(\cup_{i\in [m]}B_i)$.
\end{enumerate}

\begin{algorithm}[hptb]
\caption{Approximation Algorithm for $\textbf{P.0}$ when $\alpha\leq1/2$}
\label{alg:2}
\begin{algorithmic}[1]
\STATE Apply  $\mathcal{A}$  to solve $\textbf{P.1}$ and obtain a solution $A^{P.1}$
\FOR {every group $i\in[m]$}
\IF {$|A^{P.1} \cap V_i|< \lfloor\alpha |V_i|\rfloor$}
\STATE  select a random backup set $B_i$ of size $\lfloor\alpha |V_i|\rfloor-|A^{P.1} \cap V_i|$ from $V_i\setminus A^{P.1}$
\ELSE
\STATE $B_i\leftarrow \emptyset$
\ENDIF
\ENDFOR
\STATE $A^{approx}\leftarrow A^{P.1}\cup(\cup_{i\in [m]}B_i)$
\RETURN $A^{approx}$
\end{algorithmic}
\end{algorithm}

The pseudocode of this approximation algorithm is given as Algorithm \ref{alg:2}. Observe that   $A^{P.1}$ is a feasible solution to $\textbf{P.1}$, hence, $A^{P.1}$  satisfies upper bound constraints of $\textbf{P.1}$ and hence $\textbf{P.0}$, i.e., $|S \cap V_i| \leq \lfloor\beta |V_i|\rfloor, \forall i \in[m]$. According to the construction of $B_i$, it is easy to verify that  adding $\cup_{i\in [m]}B_i$ to $A^{P.1}$ does not violate the upper bound constraints because $\cup_{i\in [m]}B_i$ are only supplemented to those groups which do not satisfy the lower bound constraints of  $\textbf{P.0}$, i.e., $\lfloor\alpha |V_i|\rfloor \leq |S \cap V_i|$. Moreover, adding $\cup_{i\in [m]}B_i$ to $A^{P.1}$ makes it satisfy lower bound constraints of   $\textbf{P.0}$. Hence, $A^{approx}$ is a feasible solution to $\textbf{P.0}$.
 \begin{lemma}
 $A^{approx}$ is a feasible solution to $\textbf{P.0}$.
 \end{lemma}
\subsubsection{Performance Analysis}
\label{sec:performance1}
We next analyze the performance of Algorithm \ref{alg:2}. We first introduce a useful lemma from  \cite{buchbinder2014submodular}.

\begin{lemma}If $f$ is submodular and $S$ is a random subset of $V$, such that each item in $V$ is contained in
$S$ with probability at most $p$, then
$\mathbb{E}_S[f(S)] \geq (1-p)f(\emptyset)$.
\label{lem:ama}
\end{lemma}

The next lemma states that if $A^{P.1}$ is a $\gamma$-approximate solution of $\textbf{P.1}$, then $f(A^{P.1})$ is at least $\gamma$ fraction of the optimal solution of $\textbf{P.0}$.
 \begin{lemma}
 \label{lem:seq}
Suppose $\mathcal{A}$ is a $\gamma$-approximate algorithm for non-monotone submodular
maximization subject to a matroid constraint. Let $OPT$ denote the optimal solution of $\textbf{P.0}$, we have $f(A^{P.1})\geq \gamma f(OPT)$.
 \end{lemma}
\emph{Proof:} Because $\mathcal{A}$ is a $\gamma$-approximate algorithm for non-monotone submodular
maximization subject to a matroid constraint, we have $f(A^{P.1})\geq \gamma f(O^{P.1})$ where $O^{P.1}$ denotes the optimal solution of $\textbf{P.1}$. Moreover, because $\textbf{P.1}$ is a relaxed version of $\textbf{P.0}$, we have  $ f(O^{P.1})\geq f(OPT)$. Hence, $f(A^{P.1})\geq \gamma f(OPT)$. $\Box$

 We next show that augmenting $A^{P.1}$ with  items from the random set $\cup_{i\in [m]}B_i$ reduces its utility  by a factor of at most $1/2$ in expectation. Here the expectation is taken over the distribution of $\cup_{i\in [m]}B_i$.

 \begin{lemma}
 \label{lem:seq2}
Suppose  $\alpha\leq 1/2$, we have $\mathbb{E}_{A^{approx}}[f(A^{approx})]\geq \frac{1}{2}f(A^{P.1})$ where $A^{approx}= A^{P.1}\cup(\cup_{i\in [m]}B_i)$.
 \end{lemma}
\emph{Proof:} Recall that $B_i= \emptyset$ for all $i\in[m]$ such that $|A^{P.1} \cap V_i|\geq \lfloor\alpha |V_i|\rfloor$, hence, adding those $B_i$ to $A^{P.1}$ does not affect its utility. In the rest of the proof we focus on those $B_i$ with
\begin{eqnarray}
\label{eq:aamas}
|A^{P.1} \cap V_i|< \lfloor\alpha |V_i|\rfloor.
 \end{eqnarray}
 Recall that for every $i\in[m]$ such that $|A^{P.1} \cap V_i|< \lfloor\alpha |V_i|\rfloor$, $B_i$ is a random set of size $\lfloor\alpha |V_i|\rfloor-|A^{P.1} \cap V_i|$ that is sampled from $V_i\setminus A^{P.1}$. It follows that each item in  $V_i\setminus A^{P.1}$ is contained in  $B_i$  with probability at most
\begin{eqnarray}
\label{eq:haha}
\frac{\lfloor\alpha |V_i|\rfloor-|A^{P.1} \cap V_i|}{|V_i\setminus A^{P.1}|}.
\end{eqnarray}

We next give an upper bound of (\ref{eq:haha}). First,
\begin{eqnarray}
\label{eq:haha1}
\lfloor\alpha |V_i|\rfloor-|A^{P.1} \cap V_i|\leq \lfloor\alpha |V_i|\rfloor\leq  |V_i|/2,
\end{eqnarray} where the second inequality is by the assumption that $\alpha\leq 1/2$.
Moreover,
\begin{eqnarray}
&&|V_i\setminus A^{P.1}|=|V_i|- |A^{P.1} \cap V_i|=(\lfloor\alpha |V_i|\rfloor-|A^{P.1} \cap V_i|)+ (|V_i|-\lfloor\alpha |V_i|\rfloor)\\
&&\geq (\lfloor\alpha |V_i|\rfloor-|A^{P.1} \cap V_i|)+|V_i|/2,\label{eq:haha2}
\end{eqnarray}
where the inequality is by the assumption that $\alpha\leq 1/2$.

Hence,
 \begin{eqnarray}
 (\ref{eq:haha}) \leq \frac{\lfloor\alpha |V_i|\rfloor-|A^{P.1} \cap V_i|}{ (\lfloor\alpha |V_i|\rfloor-|A^{P.1} \cap V_i|)+|V_i|/2}\leq \frac{|V_i|/2}{|V_i|/2+|V_i|/2}=1/2,
  \end{eqnarray}
 where the first inequality is by  (\ref{eq:haha2}); the second inequality is by (\ref{eq:haha1}) and the assumption that $\lfloor\alpha |V_i|\rfloor-|A^{P.1} \cap V_i|>0$ (listed in (\ref{eq:aamas})).
 That is, the probability that each item  in $V_i\setminus A^{P.1}$ is contained in $B_i$ is at most $1/2$. It follows that the probability that each item in $V \setminus A^{P.1}$ is contained in $\cup_{i\in [m]}B_i$ is at most $1/2$. Moreover,
Lemma \ref{lem:ama} states that if $f$ is submodular and $S$ is a random subset of $V$, such that each item in $V$ appears in
$S$ with probability at most $p$, then
$\mathbb{E}_A[f(A)] \geq (1-p)f(\emptyset)$. With the above discussion and the observation that $f(A^{P.1}\cup\cdot)$ is submodular, it holds that $\mathbb{E}_{A^{approx}}[f(A^{approx})]=\mathbb{E}_{\cup_{i\in [m]}B_i}[f(A^{P.1}\cup(\cup_{i\in [m]}B_i))]\geq  (1-\frac{1}{2})f(A^{P.1}\cup\emptyset)=\frac{1}{2}f(A^{P.1})$.
$\Box$

Our main theorem as below follows from Lemma \ref{lem:seq} and Lemma \ref{lem:seq2}.

 \begin{theorem}
 \label{thm:1}
Suppose $\mathcal{A}$ is a $\gamma$-approximate algorithm for non-monotone submodular
maximization subject to a matroid constraint and $\alpha\leq 1/2$, we have $\mathbb{E}_{A^{approx}}[f(A^{approx})]\geq \frac{\gamma}{2}f(OPT)$.
 \end{theorem}

One option of $\mathcal{A}$  is the continuous double greedy algorithm proposed in \cite{feldman2011unified} which gives
 a $1/e-o(1)$-approximation solution, that is, $\gamma\geq 1/e-o(1)$. This, together with Theorem \ref{thm:1}, implies that $\mathbb{E}_{A^{approx}}[f(A^{approx})]\geq \frac{1/e-o(1)}{2}f(OPT)$.

\subsection{The case when $\alpha> 1/2$}
We next consider the case when  $\alpha> 1/2$. We first introduce a new utility function $g: 2^V\rightarrow \mathbb{R}_+$ as below:
\begin{eqnarray}
g(\cdot) = f(V\setminus \cdot).
\end{eqnarray} We first present a well-known result, which states that   submodular functions maintain their submodularity property when taking their complement.

\begin{lemma}
\label{lem:submodular}
If $f$ is submodular, then $g$ must be submodular.
\end{lemma}
%
%

With utility function $g$, we present a new optimization problem $\textbf{P.2}$ as below:

 \begin{center}
\framebox[0.6\textwidth][c]{
\enspace
\begin{minipage}[t]{0.6\textwidth}
\small
$\textbf{P.2}$
$\max g(S)$ \\
\textbf{subject to:}  \\
$|V_i|-\lfloor\beta |V_i|\rfloor \leq |S \cap V_i| \leq |V_i|-\lfloor\alpha |V_i|\rfloor, \forall i \in[m]$.
\end{minipage}
}
\end{center}
\vspace{0.1in}

$\textbf{P.2}$ is a flipped version of the original problem $\textbf{P.0}$ in the sense that if there is a $\gamma$-approximate solution $A^{P.2}$ to $\textbf{P.2}$, it can be easily verified that $V\setminus A^{P.2}$ is a $\gamma$-approximate solution to $\textbf{P.0}$. As a result, we will focus on solving $\textbf{P.2}$ for the rest of this section.

To solve $\textbf{P.2}$, we introduce another problem (labeled as $\textbf{P.3}$) as follows:

 \begin{center}
\framebox[0.4\textwidth][c]{
\enspace
\begin{minipage}[t]{0.4\textwidth}
\small
$\textbf{P.3}$
$\max g(S)$ \\
\textbf{subject to:}  \\
$|S \cap V_i| \leq |V_i|-\lfloor\alpha |V_i|\rfloor, \forall i \in[m]$.
\end{minipage}
}
\end{center}
\vspace{0.1in}

$\textbf{P.3}$ is relaxed version of  $\textbf{P.2}$ with lower bound constraints $|V_i|-\lfloor\beta |V_i|\rfloor \leq |S \cap V_i|$ in $\textbf{P.2}$ being removed. Because $g$ is a submodular function, $\textbf{P.3}$ is a classic problem of maximizing a submodular function subject to matroid constraints. Now we are ready to present the design of our algorithm.
\begin{enumerate}
\item Apply the state-of-the-art algorithm $\mathcal{A}$ for matroid constrained submodular maximization to solve $\textbf{P.3}$ and obtain a solution $A^{P.3}$.

\item Note that $A^{P.3}$ is not necessarily a feasible solution to $\textbf{P.2}$ because it might violate the lower bound constraints $|V_i|-\lfloor\beta |V_i|\rfloor \leq |S \cap V_i|$  for some groups. We add additional items to $A^{P.3}$ to make it feasible. Specifically, for each group $i\in[m]$ such that $|A^{P.3} \cap V_i|< |V_i|-\lfloor\beta |V_i|\rfloor$, our algorithm selects a backup set $B_i$ of size $|V_i|-\lfloor\beta |V_i|\rfloor -|A^{P.3} \cap V_i|$, by randomly sampling  $|V_i|-\lfloor\beta |V_i|\rfloor -|A^{P.3} \cap V_i|$ items from $V_i\setminus A^{P.3}$. Define $B_i=\emptyset$ if $|A^{P.1} \cap V_i|\geq |V_i|-\lfloor\beta |V_i|\rfloor$.
\item Add $\cup_{i\in [m]}B_i$ to $A^{P.3}$ to build $A^{approx}$, i.e., $A^{approx}= A^{P.3}\cup(\cup_{i\in [m]}B_i)$. Return $V\setminus A^{approx}$ as the final solution.
\end{enumerate}

\begin{algorithm}[hptb]
\caption{Approximation Algorithm for $\textbf{P.0}$ when $\alpha>1/2$}
\label{alg:3}
\begin{algorithmic}[1]
\STATE Apply  $\mathcal{A}$  to solve $\textbf{P.3}$ and obtain a solution $A^{P.3}$
\FOR {every group $i\in[m]$}
\IF {$|A^{P.3} \cap V_i|< |V_i|-\lfloor\beta |V_i|\rfloor$}
\STATE  select a random backup set $B_i$ of size $|V_i|-\lfloor\beta |V_i|\rfloor -|A^{P.3} \cap V_i|$ from $V_i\setminus A^{P.3}$
\ELSE
\STATE $B_i\leftarrow \emptyset$
\ENDIF
\ENDFOR
\STATE $A^{approx}\leftarrow A^{P.3}\cup(\cup_{i\in [m]}B_i)$
\RETURN $V\setminus A^{approx}$
\end{algorithmic}
\end{algorithm}

The pseudocode of this approximation algorithm is given as Algorithm \ref{alg:3}. Observe that $A^{P.3}$ satisfies upper bound constraints of  $\textbf{P.3}$ and hence $\textbf{P.2}$ because  $A^{P.3}$ is a feasible solution to $\textbf{P.3}$.  According to the construction of $B_i$, adding $\cup_{i\in [m]}B_i$ to $A^{P.1}$ does not violate the upper bound constraints because $\cup_{i\in [m]}B_i$ are added to meet the lower bound constraints of  $\textbf{P.2}$ if necessary.  Moreover, adding $\cup_{i\in [m]}B_i$ to $A^{P.3}$ makes it satisfy lower bound constraints of   $\textbf{P.2}$. Hence, $A^{approx}$ is a feasible solution to $\textbf{P.2}$.
 \begin{lemma}
 $A^{approx}$ is a feasible solution to $\textbf{P.2}$.
 \end{lemma}

\subsubsection{Performance Analysis}
\label{sec:performance2}
 We first introduce a technical lemma which states that if $A^{P.3}$ is a $\gamma$-approximate solution of $\textbf{P.3}$, then $f(A^{P.3})$ is at least $\gamma$ fraction of the optimal solution of $\textbf{P.2}$. This lemma follows from the observation that $\textbf{P.3}$ is a relaxation of  $\textbf{P.2}$ .
 \begin{lemma}
 \label{lem:jian}
Suppose $\mathcal{A}$ is a $\gamma$-approximate algorithm for non-monotone submodular
maximization subject to a matroid constraint. Let $O^{P.2}$ denote the optimal solution of $\textbf{P.2}$, it holds that $g(A^{P.3})\geq \gamma g(O^{P.2})$.
 \end{lemma}

 We next show that augmenting $A^{P.3}$ with  items from $\cup_{i\in [m]}B_i$ reduces its utility  by a factor of at most $2/3$ in expectation.

 \begin{lemma}
 \label{lem:ti}
Suppose  $\alpha> 1/2$, $\mathbb{E}_{A^{approx}}[g(A^{approx})]\geq \frac{1}{3}g(A^{P.3})$ where $A^{approx}= A^{P.3}\cup(\cup_{i\in [m]}B_i)$.
 \end{lemma}
\emph{Proof:} Recall that $B_i= \emptyset$ for all $i\in[m]$ such that $|A^{P.3} \cap V_i|\geq |V_i|-\lfloor\beta |V_i|\rfloor$, hence, adding those $B_i$ to $A^{P.3}$ does not affect its utility. Therefore, we focus on those groups $i\in[m]$ with $|A^{P.3} \cap V_i|< |V_i|-\lfloor\beta |V_i|\rfloor$ in the rest of the proof. Let $M=\{i\mid |A^{P.3} \cap V_i|< |V_i|-\lfloor\beta |V_i|\rfloor\}$ denote the set containing the indexes of all such groups and we assume $M\neq \emptyset$ to avoid trivial cases. We next show that it is safe to assume $\min_{i\in M}|V_i|>1$ without loss of generality, i.e., the smallest group in $M$ contains at least two items. To prove this, we consider two cases, depending on the value of $\beta$. If $\beta=1$, then $|A^{P.3} \cap V_i|< |V_i|-\lfloor\beta |V_i|\rfloor$ does not hold for any group $i$ such that $|V_i|=1$, that is, $\min_{i\in M}|V_i|>1$. If $\beta<1$, then according to the group fairness constraints listed in  $\textbf{P.0}$, we are not allowed to select any items from those groups with $|V_i|=1$. Hence, removing all groups with size one from consideration does not affect  the quality of the optimal solution.

With the assumption that $\min_{i\in M}|V_i|>1$, we are now in position to prove this lemma.  Recall that for every $i\in M$, $B_i$ is a random set of size $|V_i|-\lfloor\beta |V_i|\rfloor -|A^{P.3} \cap V_i|$ that is sampled from $V_i\setminus A^{P.3}$. It follows that each item in  $V_i\setminus A^{P.3}$ appears in $B_i$ with probability at most
\begin{eqnarray}
\label{eq:haha3}
\frac{|V_i|-\lfloor\beta |V_i|\rfloor -|A^{P.3} \cap V_i|}{|V_i\setminus A^{P.3}|}.
\end{eqnarray}

We next give an upper bound of (\ref{eq:haha3}). Because we assume $\alpha> 1/2$, we have $\beta\geq\alpha>1/2$. This, together with the assumption that $\min_{i\in M}|V_i|>1$, implies that for all $i\in M$,
\begin{eqnarray}
\label{eq:haha4}
|V_i|-\lfloor\beta |V_i|\rfloor -|A^{P.3} \cap V_i| \leq  |V_i|-\lfloor\beta |V_i|\rfloor \leq 2|V_i|/3.
\end{eqnarray}
Moreover,
\begin{eqnarray}
&&|V_i\setminus A^{P.3}|=|V_i|-|A^{P.3} \cap V_i|\\
&&=(|V_i|-\lfloor\beta |V_i|\rfloor -|A^{P.3} \cap V_i|)+(|V_i|-(|V_i|-\lfloor\beta |V_i|\rfloor)) \\
&&= (|V_i|-\lfloor\beta |V_i|\rfloor -|A^{P.3} \cap V_i|)+\lfloor\beta |V_i|\rfloor \\
&&\geq (|V_i|-\lfloor\beta |V_i|\rfloor -|A^{P.3} \cap V_i|)+|V_i|/3, \label{eq:haha5}
\end{eqnarray}
where the inequality is by the observation that $\beta>1/2$. It follows that
\begin{eqnarray}
(\ref{eq:haha3})\leq \frac{|V_i|-\lfloor\beta |V_i|\rfloor -|A^{P.3} \cap V_i|}{(|V_i|-\lfloor\beta |V_i|\rfloor -|A^{P.3} \cap V_i|)+|V_i|/3}\leq\frac{2|V_i|/3}{2|V_i|/3+|V_i|/3}=2/3,
\end{eqnarray}
where the first inequality is by (\ref{eq:haha5}) and the second inequality is by (\ref{eq:haha4}) and the assumption that $|V_i|-\lfloor\beta |V_i|\rfloor -|A^{P.3} \cap V_i|>0$. That is, each item in $V_i\setminus A^{P.3}$ appears in  $B_i$ with probability at most $2/3$. Lemma \ref{lem:ama} and the observation that $g(A^{P.3}\cup\cdot)$ is submodular imply that $\mathbb{E}_{A^{approx}}[g(A^{approx})]=\mathbb{E}_{\cup_{i\in [m]}B_i}[g(A^{P.3}\cup(\cup_{i\in [m]}B_i))]\geq (1-\frac{2}{3})g(A^{P.3}\cup\emptyset)=  \frac{1}{3}g(A^{P.3})$. $\Box$

Lemma \ref{lem:jian} and Lemma \ref{lem:ti} together imply that \[\mathbb{E}_{A^{approx}}[g(A^{approx})]\geq \frac{1}{3}g(A^{P.3})\geq  \frac{\gamma}{3}g(O^{P.2}).\] By the definition of function $g$, we have  \[\mathbb{E}_{A^{approx}}[f(V\setminus A^{approx})]=\mathbb{E}_{A^{approx}}[g(A^{approx})] \geq  \frac{\gamma}{3}g(O^{P.2})=\frac{\gamma}{3}f(OPT)\] where the last equality is by the observation that $\textbf{P.2}$ and $\textbf{P.0}$ share the same value of the optimal solution. Hence, the following main theorem holds.

 \begin{theorem}
 \label{thm:2}
Suppose $\mathcal{A}$ is a $\gamma$-approximate algorithm for non-monotone submodular
maximization subject to a matroid constraint and $\alpha> 1/2$, we have $\mathbb{E}_{A^{approx}}[f(V\setminus A^{approx})]\geq \frac{\gamma}{3}f(OPT)$.
 \end{theorem}

If we adopt the continuous double greedy algorithm  \cite{feldman2011unified} as $\mathcal{A}$ to compute  $A^{P.3}$,  it gives
 a $1/e-o(1)$-approximation solution, that is, $\gamma\geq 1/e-o(1)$. This, together with Theorem \ref{thm:2}, implies that $\mathbb{E}_{A^{approx}}[f(V\setminus A^{approx})]\geq \frac{1/e-o(1)}{3}f(OPT)$.
\section{Extension: Incorporating Global Cardinality Constraint}
\label{sec:non-adaptive-extended}

In this section, we extend $\textbf{P.0}$ to incorporate a global cardinality constraint. A formal definition of this problem is listed in $\textbf{P.A}$. Our objective is to find a best $S$ subject to a group fairness constraint $(\alpha,\beta)$ and an additional cardinality constraint $c$.

 \begin{center}
\framebox[0.6\textwidth][c]{
\enspace
\begin{minipage}[t]{0.6\textwidth}
\small
$\textbf{P.A}$
$\max f(S)$ \\
\textbf{subject to:}  \\
$\lfloor\alpha |V_i|\rfloor \leq |S \cap V_i| \leq \lfloor\beta |V_i|\rfloor, \forall i \in[m]$.\\
$|S|\leq c$.
\end{minipage}
}
\end{center}
\vspace{0.1in}

\subsection{The case when $\alpha\leq 1/2$}
We first consider the case when $\alpha\leq 1/2$. We introduce a new optimization problem $\textbf{P.B}$ as follows:
 \begin{center}
\framebox[0.6\textwidth][c]{
\enspace
\begin{minipage}[t]{0.6\textwidth}
\small
$\textbf{P.B}$
$\max f(S)$ \\
\textbf{subject to:}  \\
$|S \cap V_i| \leq \lfloor\beta |V_i|\rfloor, \forall i \in[m]$.\\
$\sum_{i\in[m]}\max\{\lfloor\alpha |V_i|\rfloor , |S\cap V_i|\} \leq c$.
\end{minipage}
}
\end{center}
\vspace{0.1in}

It is easy to verify that $\textbf{P.B}$ is a relaxation of $\textbf{P.A}$ in the sense that every feasible solution to $\textbf{P.A}$ is also a feasible solution to  $\textbf{P.B}$. Hence, we have the following lemma.

 \begin{lemma}
 \label{lem:ti1}
Let $OPT$ denote the optimal solution of $\textbf{P.A}$ and $O^{P.B}$ denote the optimal solution of $\textbf{P.B}$, we have $f(O^{P.B})\geq f(OPT)$.
 \end{lemma}

It has been shown that the constraints in $\textbf{P.B}$ gives rise to a matroid \cite{el2020fairness}. This, together with the assumption that $f$ is a submodular function, implies that $\textbf{P.B}$ is a classic problem of maximizing a submodular function subject to  matroid constraints. Now we are ready to present the design of our algorithm.
\begin{enumerate}
\item Apply the state-of-the-art algorithm $\mathcal{A}$ for matroid constrained submodular maximization to solve $\textbf{P.B}$ and obtain a solution $A^{P.B}$.

\item Note that $A^{P.B}$ is not necessarily a feasible solution to $\textbf{P.A}$ because it might violate the lower bound constraints $\lfloor\alpha |V_i|\rfloor \leq |S \cap V_i|$ for some groups. To make it feasible, we add additional items to $A^{P.B}$. Specifically, for each group $i\in[m]$ such that $|A^{P.B} \cap V_i|< \lfloor\alpha |V_i|\rfloor$, our algorithm selects a backup set $B_i$ of size $\lfloor\alpha |V_i|\rfloor-|A^{P.B} \cap V_i|$, by randomly sampling  $\lfloor\alpha |V_i|\rfloor-|A^{P.B} \cap V_i|$ items from $V_i\setminus A^{P.B}$. Define $B_i=\emptyset$ if $|A^{P.1} \cap V_i|\geq \lfloor\alpha |V_i|\rfloor$.
\item  At the end, add $\cup_{i\in [m]}B_i$ to $A^{P.B}$ to build the final solution $A^{approx}$, i.e., $A^{approx}= A^{P.B}\cup(\cup_{i\in [m]}B_i)$.
\end{enumerate}

\begin{algorithm}[hptb]
\caption{Approximation Algorithm for $\textbf{P.A}$ when $\alpha\leq 1/2$}
\label{alg:21}
\begin{algorithmic}[1]
\STATE Apply  $\mathcal{A}$  to solve $\textbf{P.B}$ and obtain a solution $A^{P.B}$
\FOR {every group $i\in[m]$}
\IF {$|A^{P.B} \cap V_i|< \lfloor\alpha |V_i|\rfloor$}
\STATE  select a random backup set $B_i$ of size $\lfloor\alpha |V_i|\rfloor-|A^{P.B} \cap V_i|$ from $V_i\setminus A^{P.B}$
\ELSE
\STATE $B_i\leftarrow \emptyset$
\ENDIF
\ENDFOR
\STATE $A^{approx}\leftarrow A^{P.B}\cup(\cup_{i\in [m]}B_i)$
\RETURN $A^{approx}$
\end{algorithmic}
\end{algorithm}

The pseudocode of this approximation algorithm is given as Algorithm \ref{alg:21}. Observe that $A^{P.B}$ satisfies the group-wise upper bound constraints of  $\textbf{P.A}$ because  $A^{P.B}$ meets the first set of constraints in $\textbf{P.B}$. According to the construction of $B_i$, adding $\cup_{i\in [m]}B_i$ to $A^{P.B}$ does not violate the group-wise upper bound constraints of  $\textbf{P.A}$ because $\cup_{i\in [m]}B_i$ are added to meet the lower bound constraints of  $\textbf{P.A}$ if necessary.   Moreover, adding $\cup_{i\in [m]}B_i$ to $A^{P.B}$ does not violate the global cardinality constraint of  $\textbf{P.A}$ because $A^{P.B}$ meets the second set of constraints in $\textbf{P.B}$. At last, it is easy to verify that adding $\cup_{i\in [m]}B_i$ to $A^{P.B}$ makes it satisfy the lower bound constraints of   $\textbf{P.A}$. Hence, $A^{approx}$ is a feasible solution to $\textbf{P.A}$.
 \begin{lemma}
 $A^{approx}$ is a feasible solution to $\textbf{P.A}$.
 \end{lemma}

Following the same proof of Theorem \ref{thm:1}, we have the following theorem.
 \begin{theorem}
 \label{thm:11}
Suppose $\mathcal{A}$ is a $\gamma$-approximate algorithm for non-monotone submodular
maximization subject to a matroid constraint and $\alpha\leq 1/2$, we have $\mathbb{E}_{A^{approx}}[f(A^{approx})]\geq \frac{\gamma}{2}f(OPT)$.
 \end{theorem}

 \subsection{The case when $\alpha> 1/2$}
We next consider the case when  $\alpha> 1/2$. Recall that $g(\cdot) = f(V\setminus \cdot)$.  We first present a flipped formation of $\textbf{P.A}$ as below:

 \begin{center}
\framebox[0.6\textwidth][c]{
\enspace
\begin{minipage}[t]{0.6\textwidth}
\small
$\textbf{P.C}$
$\max g(S)$ \\
\textbf{subject to:}  \\
$|V_i|-\lfloor\beta |V_i|\rfloor \leq |S \cap V_i| \leq |V_i|-\lfloor\alpha |V_i|\rfloor, \forall i \in[m]$.\\
$|S|\geq n-c$.
\end{minipage}
}
\end{center}
\vspace{0.1in}

Suppose there is a $\gamma$-approximate solution $A^{P.C}$ to $\textbf{P.C}$, it is easy to verify that $V\setminus A^{P.C}$ is a $\gamma$-approximate solution to $\textbf{P.A}$. We focus on solving $\textbf{P.C}$ in the rest of this section. We first introduce a new optimization problem (labeled as $\textbf{P.D}$) as follows:

 \begin{center}
\framebox[0.4\textwidth][c]{
\enspace
\begin{minipage}[t]{0.4\textwidth}
\small
$\textbf{P.D}$
$\max g(S)$ \\
\textbf{subject to:}  \\
$|S \cap V_i| \leq |V_i|-\lfloor\alpha |V_i|\rfloor, \forall i \in[m]$.
\end{minipage}
}
\end{center}
\vspace{0.1in}

$\textbf{P.D}$ is relaxed version of  $\textbf{P.C}$ with both group-wise lower bound constraints $|V_i|-\lfloor\beta |V_i|\rfloor \leq |S \cap V_i|$ and global  lower bound constraints $|S|\geq n-c$ in $\textbf{P.C}$ being removed. Hence, we have the following lemma.

 \begin{lemma}
 \label{lem:ti2}
Let $O^{P.C}$ denote the optimal solution of $\textbf{P.C}$  and $O^{P.D}$ denote the optimal solution of $\textbf{P.D}$, we have $g(O^{P.D})\geq g(O^{P.C})$.
 \end{lemma}

Recall that if $f$ is submodular, $g$ must be submodular (by Lemma \ref{lem:submodular}). Hence, $\textbf{P.D}$ is a classic problem of maximizing a submodular function subject to matroid constraints. We next present the design of our algorithm.
\begin{enumerate}
\item Apply the state-of-the-art algorithm $\mathcal{A}$ for matroid constrained submodular maximization to solve $\textbf{P.D}$ and obtain a solution $A^{P.D}$.

\item Note that $A^{P.D}$ is not necessarily a feasible solution to $\textbf{P.C}$ because it might violate the group-wise or the global lower bound constraints of  $\textbf{P.C}$. We add additional items to $A^{P.D}$ to make it feasible. Specifically, for each group $i\in[m]$, our algorithm selects a backup set $B_i$ of size $|V_i|-\lfloor\alpha |V_i|\rfloor -|A^{P.D} \cap V_i|$, by randomly sampling  $|V_i|-\lfloor\alpha |V_i|\rfloor -|A^{P.D} \cap V_i|$ items from $V_i\setminus A^{P.D}$. Define $B_i=\emptyset$ if $|V_i|-\lfloor\alpha |V_i|\rfloor -|A^{P.D} \cap V_i|=0$.
\item Add $\cup_{i\in [m]}B_i$ to $A^{P.D}$ to build  $A^{approx}$, i.e., $A^{approx}= A^{P.D}\cup(\cup_{i\in [m]}B_i)$. Return $V\setminus A^{approx}$ as the final solution.
\end{enumerate}

\begin{algorithm}[hptb]
\caption{Approximation Algorithm for $\textbf{P.A}$ when $\alpha>1/2$}
\label{alg:31}
\begin{algorithmic}[1]
\STATE Apply  $\mathcal{A}$  to solve $\textbf{P.D}$ and obtain a solution $A^{P.D}$
\FOR {every group $i\in[m]$}
\IF {$|A^{P.D} \cap V_i|< |V_i|-\lfloor\alpha |V_i|\rfloor$}
\STATE  select a random backup set $B_i$ of size  $|V_i|-\lfloor\alpha |V_i|\rfloor -|A^{P.D} \cap V_i|$  from $V_i\setminus A^{P.D}$
\ELSE
\STATE $B_i\leftarrow \emptyset$
\ENDIF
\ENDFOR
\STATE $A^{approx}\leftarrow A^{P.D}\cup(\cup_{i\in [m]}B_i)$
\RETURN $V\setminus A^{approx}$
\end{algorithmic}
\end{algorithm}

The pseudocode of this approximation algorithm is given as Algorithm \ref{alg:31}. Observe that adding $\cup_{i\in [m]}B_i$ to $A^{P.D}$ ensures that each group contributes exactly $|V_i|-\lfloor\alpha |V_i|\rfloor$ number of items to the solution. Because $n-c\leq \sum_{i\in[m]}(|V_i|-\lfloor\alpha |V_i|\rfloor)$ (otherwise $\textbf{P.C}$ does not have a feasible solution), $A^{P.D}\cup(\cup_{i\in [m]}B_i)$ must satisfy all constraints in  $\textbf{P.C}$. Hence, we have the following lemma.
 \begin{lemma}
 $A^{approx}$ is a feasible solution to $\textbf{P.C}$.
 \end{lemma}

We next analyze the performance of $A^{approx}$. The following lemma states that adding $\cup_{i\in [m]}B_i$ to  $A^{P.D}$ reduces its utility  by a factor of at most $2/3$ in expectation.

 \begin{lemma}
 \label{lem:ti3}
Suppose  $\alpha> 1/2$, we have $\mathbb{E}_{A^{approx}}[g(A^{approx})]\geq \frac{1}{3}g(A^{P.D})$.
 \end{lemma}
\emph{Proof:} Observe that $B_i= \emptyset$ for all $i\in[m]$ such that $|A^{P.D} \cap V_i|= |V_i|-\lfloor\alpha |V_i|\rfloor$, hence, adding those $B_i$ to $A^{P.D}$ does not affect its utility. Therefore, we focus on those groups $i\in[m]$ with $|A^{P.D} \cap V_i|< |V_i|-\lfloor\alpha |V_i|\rfloor$ in the rest of the proof. Let $Z=\{i\in[m]\mid |A^{P.D} \cap V_i|< |V_i|-\lfloor\alpha |V_i|\rfloor\}$ denote the set containing the indexes all such groups. We assume $Z\neq \emptyset$ to avoid trivial cases.  We next show that it is safe to assume $\min_{i\in Z}|V_i|>1$ without loss of generality, i.e., the smallest group in $Z$ contains at least two items. To prove this, we consider two cases, depending on the value of $\alpha$. If $\alpha=1$, then $|A^{P.D} \cap V_i|< |V_i|-\lfloor\alpha |V_i|\rfloor$ does not hold for any group $i$ such that $|V_i|=1$. Hence, $\min_{i\in Z}|V_i|>1$. If $\alpha<1$, then according to the group fairness constraints listed in  $\textbf{P.A}$, we are not allowed to select any items from those groups with $|V_i|=1$. Hence, removing all groups with size one from consideration does not affect the quality of the optimal solution.

With the assumption that $\min_{i\in Z}|V_i|>1$, we are now ready to prove this lemma.  Recall that for every $i\in Z$, $B_i$ is a random set of size $|V_i|-\lfloor\alpha |V_i|\rfloor -|A^{P.D} \cap V_i|$ that is sampled from $V_i\setminus A^{P.D}$. It follows each item in  $V_i\setminus A^{P.D}$ appears in  $B_i$  with probability at most
\begin{eqnarray}
\label{eq:haha31}
\frac{|V_i|-\lfloor\alpha |V_i|\rfloor -|A^{P.D} \cap V_i|}{|V_i\setminus A^{P.D}|}.
\end{eqnarray}

We next give an upper bound of (\ref{eq:haha31}). Because we assume $\alpha> 1/2$ and $\min_{i\in Z}|V_i|>1$, it holds that for all $i\in M$,
\begin{eqnarray}
\label{eq:haha41}
|V_i|-\lfloor\alpha |V_i|\rfloor -|A^{P.D} \cap V_i| \leq  |V_i|-\lfloor\alpha |V_i|\rfloor \leq 2|V_i|/3.
\end{eqnarray}
Moreover,
\begin{eqnarray}
&&|V_i\setminus A^{P.D}|= |V_i|-|A^{P.D} \cap V_i|\\
&&=(|V_i|-\lfloor\alpha |V_i|\rfloor -|A^{P.D} \cap V_i|)+(|V_i|-(|V_i|-\lfloor\alpha |V_i|\rfloor)) \\
&&= (|V_i|-\lfloor\alpha |V_i|\rfloor -|A^{P.D} \cap V_i|)+\lfloor\alpha |V_i|\rfloor \\
&&\geq (|V_i|-\lfloor\alpha |V_i|\rfloor -|A^{P.D} \cap V_i|)+|V_i|/3, \label{eq:haha51}
\end{eqnarray}
where the inequality is by the assumptions that $\alpha>1/2$ and $|V_i|>1$. It follows that
\begin{eqnarray}
(\ref{eq:haha31})\leq \frac{|V_i|-\lfloor\alpha |V_i|\rfloor -|A^{P.D} \cap V_i|}{(|V_i|-\lfloor\alpha |V_i|\rfloor -|A^{P.D} \cap V_i|)+|V_i|/3}\leq\frac{2|V_i|/3}{2|V_i|/3+|V_i|/3}=2/3,
\end{eqnarray}
where the first inequality is by (\ref{eq:haha51}) and the second inequality is by (\ref{eq:haha41}) and the assumption that $|V_i|-\lfloor\alpha |V_i|\rfloor -|A^{P.D} \cap V_i|>0$. That is, each item in  $V_i\setminus A^{P.D}$ appears in $B_i$ with probability at most $2/3$. Lemma \ref{lem:ama} and the observation that $g(A^{P.D}\cup\cdot)$ is submodular imply that $\mathbb{E}_{A^{approx}}[g(A^{approx})]=\mathbb{E}_{\cup_{i\in [m]}B_i}[g(A^{P.D}\cup(\cup_{i\in [m]}B_i))]\geq (1-\frac{2}{3})g(A^{P.D}\cup\emptyset)=\frac{1}{3}g(A^{P.D})$. $\Box$

Suppose  $\mathcal{A}$ is a $\gamma$-approximate algorithm for non-monotone submodular
maximization subject to a matroid constraint, we have  \[\mathbb{E}_{A^{approx}}[g(A^{approx})]\geq \frac{1}{3}g(A^{P.D})\geq \frac{\gamma}{3}g(O^{P.D})\] where the first inequality is by Lemma \ref{lem:ti3}. This, together with $g(O^{P.D})\geq g(O^{P.C})$ (as proved in Lemma \ref{lem:ti2}), implies that  $\mathbb{E}_{A^{approx}}[g(A^{approx})]\geq \frac{\gamma}{3}g(O^{P.C})$. By the definition of function $g$, we have \[\mathbb{E}_{A^{approx}}[f(V\setminus A^{approx})] = \mathbb{E}_{A^{approx}}[g(A^{approx})]\geq \frac{\gamma}{3}g(O^{P.C})=\frac{\gamma}{3}f(OPT)\] where the last equality is by the observation that $\textbf{P.A}$ and $\textbf{P.C}$ share the same value of the optimal solution. Hence, the following main theorem holds.

 \begin{theorem}
Suppose $\mathcal{A}$ is a $\gamma$-approximate algorithm for non-monotone submodular
maximization subject to a matroid constraint and $\alpha> 1/2$, we have $\mathbb{E}_{A^{approx}}[f(V\setminus A^{approx})]\geq \frac{\gamma}{3}f(OPT)$.
 \end{theorem}

\section{Conclusion}
This paper presents a comprehensive investigation of the non-monotone submodular maximization problem under group fairness constraints. Our main contribution is the development of several constant-factor approximation algorithms for this problem. In the future, we plan to expand our research to explore alternative fairness metrics.

\bibliographystyle{splncs04}
\bibliography{reference}

\begin{thebibliography}{10}
\providecommand{\url}[1]{\texttt{#1}}
\providecommand{\urlprefix}{URL }
\providecommand{\doi}[1]{https://doi.org/#1}

\bibitem{biddle2017adverse}
Biddle, D.: Adverse impact and test validation: A practitioner's guide to valid
  and defensible employment testing. Routledge (2017)

\bibitem{buchbinder2014submodular}
Buchbinder, N., Feldman, M., Naor, J., Schwartz, R.: Submodular maximization
  with cardinality constraints. In: Proceedings of the twenty-fifth annual
  ACM-SIAM symposium on Discrete algorithms. pp. 1433--1452. SIAM (2014)

\bibitem{celis2018fair}
Celis, E., Keswani, V., Straszak, D., Deshpande, A., Kathuria, T., Vishnoi, N.:
  Fair and diverse dpp-based data summarization. In: International Conference
  on Machine Learning. pp. 716--725. PMLR (2018)

\bibitem{celis2018multiwinner}
Celis, L.E., Huang, L., Vishnoi, N.K.: Multiwinner voting with fairness
  constraints. In: Proceedings of the 27th International Joint Conference on
  Artificial Intelligence. pp. 144--151 (2018)

\bibitem{chierichetti2019matroids}
Chierichetti, F., Kumar, R., Lattanzi, S., Vassilvtiskii, S.: Matroids,
  matchings, and fairness. In: The 22nd International Conference on Artificial
  Intelligence and Statistics. pp. 2212--2220. PMLR (2019)

\bibitem{das2008algorithms}
Das, A., Kempe, D.: Algorithms for subset selection in linear regression. In:
  Proceedings of the fortieth annual ACM symposium on Theory of computing. pp.
  45--54 (2008)

\bibitem{dueck2007non}
Dueck, D., Frey, B.J.: Non-metric affinity propagation for unsupervised image
  categorization. In: 2007 IEEE 11th International Conference on Computer
  Vision. pp.~1--8. IEEE (2007)

\bibitem{10.1145/2090236.2090255}
Dwork, C., Hardt, M., Pitassi, T., Reingold, O., Zemel, R.: Fairness through
  awareness. In: Proceedings of the 3rd Innovations in Theoretical Computer
  Science Conference (2012)

\bibitem{el2011beyond}
El-Arini, K., Guestrin, C.: Beyond keyword search: discovering relevant
  scientific literature. In: Proceedings of the 17th ACM SIGKDD international
  conference on Knowledge discovery and data mining. pp. 439--447 (2011)

\bibitem{el2020fairness}
El~Halabi, M., Mitrovi{\'c}, S., Norouzi-Fard, A., Tardos, J., Tarnawski, J.M.:
  Fairness in streaming submodular maximization: algorithms and hardness.
  Advances in Neural Information Processing Systems  \textbf{33},  13609--13622
  (2020)

\bibitem{feldman2011unified}
Feldman, M., Naor, J., Schwartz, R.: A unified continuous greedy algorithm for
  submodular maximization. In: 2011 IEEE 52nd Annual Symposium on Foundations
  of Computer Science. pp. 570--579. IEEE (2011)

\bibitem{golovin2011adaptive}
Golovin, D., Krause, A.: Adaptive submodularity: Theory and applications in
  active learning and stochastic optimization. Journal of Artificial
  Intelligence Research  \textbf{42},  427--486 (2011)

\bibitem{gotovos2015non}
Gotovos, A., Karbasi, A., Krause, A.: Non-monotone adaptive submodular
  maximization. In: Twenty-Fourth International Joint Conference on Artificial
  Intelligence (2015)

\bibitem{gu2022binary}
Gu, S., Gao, C., Wu, W.: A binary search double greedy algorithm for
  non-monotone dr-submodular maximization. In: Algorithmic Aspects in
  Information and Management: 16th International Conference, AAIM 2022,
  Guangzhou, China, August 13--14, 2022, Proceedings. pp. 3--14. Springer
  (2022)

\bibitem{joseph2016fairness}
Joseph, M., Kearns, M., Morgenstern, J.H., Roth, A.: Fairness in learning:
  Classic and contextual bandits. Advances in neural information processing
  systems  \textbf{29} (2016)

\bibitem{mirzasoleiman2016fast}
Mirzasoleiman, B., Badanidiyuru, A., Karbasi, A.: Fast constrained submodular
  maximization: Personalized data summarization. In: ICML. pp. 1358--1367
  (2016)

\bibitem{monroe1995fully}
Monroe, B.L.: Fully proportional representation. American Political Science
  Review  \textbf{89}(4),  925--940 (1995)

\bibitem{shi2021k}
Shi, G., Gu, S., Wu, W.: k-submodular maximization with two kinds of
  constraints. Discrete Mathematics, Algorithms and Applications
  \textbf{13}(04),  2150036 (2021)

\bibitem{sipos2012temporal}
Sipos, R., Swaminathan, A., Shivaswamy, P., Joachims, T.: Temporal corpus
  summarization using submodular word coverage. In: Proceedings of the 21st ACM
  international conference on Information and knowledge management. pp.
  754--763 (2012)

\bibitem{tang2020influence}
Tang, S., Yuan, J.: Influence maximization with partial feedback. Operations
  Research Letters  \textbf{48}(1),  24--28 (2020)

\bibitem{tang2021adaptive}
Tang, S., Yuan, J.: Adaptive regularized submodular maximization. In: 32nd
  International Symposium on Algorithms and Computation (ISAAC 2021). Schloss
  Dagstuhl-Leibniz-Zentrum f{\"u}r Informatik (2021)

\bibitem{tang2022group}
Tang, S., Yuan, J.: Group equility in adaptive submodular maximization. arXiv
  preprint arXiv:2207.03364  (2022)

\bibitem{tang2023unified}
Tang, S., Yuan, J.: Beyond submodularity: A unified framework of fair subset
  selection through randomization. In: Under Review (2023)

\bibitem{tang2023randomization}
Tang, S., Yuan, J., Mensah-Boateng, T.: Achieving long-term fairness in
  submodular maximization through randomization. In: Under Review (2023)

\bibitem{tsang2019group}
Tsang, A., Wilder, B., Rice, E., Tambe, M., Zick, Y.: Group-fairness in
  influence maximization. arXiv preprint arXiv:1903.00967  (2019)

\bibitem{zafar2017fairness}
Zafar, M.B., Valera, I., Rogriguez, M.G., Gummadi, K.P.: Fairness constraints:
  Mechanisms for fair classification. In: Artificial intelligence and
  statistics. pp. 962--970. PMLR (2017)

\end{thebibliography}




\end{document}